\documentclass[conference]{IEEEtran}
\IEEEoverridecommandlockouts
\usepackage{cite}
\usepackage{amsmath,amssymb,amsfonts}
\usepackage{algorithmic}
\usepackage{algorithm}
\usepackage{graphicx}
\usepackage{textcomp}
\usepackage{xcolor}
\def\BibTeX{{\rm B\kern-.05em{\sc i\kern-.025em b}\kern-.08em
    T\kern-.1667em\lower.7ex\hbox{E}\kern-.125emX}}
\begin{document}
\title{Reservoir Computing Approach for Gray Images Segmentation
}

\author{\IEEEauthorblockN{Petia Koprinkova-Hristova}
\IEEEauthorblockA{\textit{Institute of Information and Communication Technologies} \\
\textit{Bulgarian Academy of Sciences}\\
Sofia, Bulgaria\\
ORCID 0000-0002-0447-9667}
}

\maketitle

\begin{abstract}
The paper proposes a novel approach for gray scale images segmentation. It is based on multiple features extraction from single feature per image pixel, namely its intensity value, using Echo state network. The newly extracted features - reservoir equilibrium states - reveal hidden image characteristics that improve its segmentation via a clustering algorithm. Moreover, it was demonstrated that the intrinsic plasticity tuning of reservoir fits its equilibrium states to the original image intensity distribution thus allowing for its better segmentation. The proposed approach is tested on the benchmark image Lena.
\end{abstract}

\begin{IEEEkeywords}
reservoir computing, Echo state network, intrinsic plasticity, image segmentation
\end{IEEEkeywords}

\section{Introduction}
Echo state networks (ESN) belong to a novel and rapidly developing family of reservoir computing approaches \cite{Jaeger,Gallicchio,LucJaeger} whose primary aim was development of fast trainable recurrent neural network (RNN) architectures able to approximate nonlinear time series dependencies.  

Following different view point to dynamic reservoir structure and its properties, in \cite{KopTontchev} a novel approach for features extraction from multidimensional data sets using ESN was proposed. It was successfully tested on numerous practical examples, among which clustering and segmentation of multi-spectral images \cite{KopCh}. Similar works dealing with ESN features extraction for image segmentation were reported in \cite{Meftah,Souahlia2020,Souahlia2017,Souahlia2016}. Other works propose to train ESN to classify image pixels based on their preliminary extracted features \cite{Jayakumari,Khan20171,Khan20172,Khan2014,Nidhi2018,Nidhi2016}. All these works are focused on colour image segmentation having multiple features per pixel, e.g. RGB grades. The gray scale images however have a single feature per pixel that is its intensity.   

The aim of the presented here investigation is to adapt the reservoir approach from \cite{KopTontchev} to uni-dimensional features data such as gray scale images and to test whether such an approach could enhance gray images segmentation via clustering. The core of the approach in \cite{KopTontchev} was to use the reservoir equilibrium states corresponding to each one of the multidimensional feature vectors as new, hopefully better feature vector of different dimension. Fitting the ESN to reflect the input data structure was achieved using an approach for tuning of reservoir internal connectivity called Intrinsic Plasticity (IP) \cite{Schrauwen,Steil}. 

In present work the IP tuned ESN reservoir with one dimensional input (pixel intensity) was exploited to extract multiple features per pixel (reservoir equilibrium states) of a gray image. The approach was investigated in details on a famous image called "Lena" that was benchmark subject in numerous image processing and analyses techniques. The original features and ESN extracted ones before and after IP tuning were compared with respect to their distribution as well as with respect of the quality of image segmentation via clustering by several approaches applied by far to Lena image, namely kmeans, fuzzy c-means, subtractive clustering \cite{Yager}, hard thresholding and  multi-level image thresholding using Otsu's method \cite{Otsu}.

The paper is organized as follows: the next section introduces ESN, IP tuning algorithm and adapted to gray scale images approach for features extraction; the following section presents obtained features of Lena image and the outcome of different clustering approaches; the paper finishes with concluding remarks and direction for future work. 

\section{ESN approach for gray scale images features extraction}
\label{ESNmethod}
\begin{figure}[H]
\centering
\includegraphics[width=0.35\textwidth]{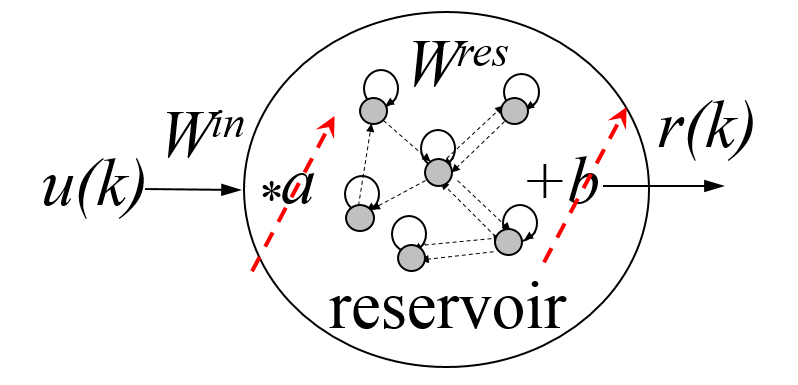}
\caption{ESN reservoir with parameters for IP tuning.}
\label{ESNeq}       
\end{figure}

The structure of an ESN reservoir is shown on Fig~\ref{ESNeq}. 

It incorporates a pool of neurons with sigmoid activation function $f^{res}$ (usually the hyperbolic tangent) that has randomly generated recurrent connection weights. The reservoir state $r(k)$ for the current time instant $k$ depends both on its previous state $r(k-1)$ and the current input $u(k)$ as follows:

\begin{equation}
     r(k)=f^{res}(diag(a)W^{in}u(k)+W^{res}r(k-1)+diag(b))
     \label{eqESNr}
\end{equation}

Here $W^{in}$ and $W^{res}$ are input and recurrent connection weight matrices that are randomly generated according to recipes given by \cite{LucJaeger}; $a$ and $b$ are vectors called gain and bias that are set to $1$ and $0$ respectively in most applications. 

In order to adjust the reservoir to the structure of its input data, \cite{Steil,Schrauwen} proposed an approach called Intrinsic Plasticity (IP) tuning that achieves desired distribution of reservoir output via changes in the gain $a$ and bias $b$ in equation (\ref{eqESNr}). The procedure is gradient algorithm minimizing Kullback-Leibler divergence between actual and target distributions. In case of hyperbolic tangent activation function the proper target distribution is normal (Gaussian) with mean $\mu$ and variance $\sigma$:

\begin{equation}
     p_{norm}=(1/{\sigma \sqrt{2\pi}})e^{(r-\mu)^2/2{\sigma}^2)}
     \label{eqGauss}
\end{equation}

Thus the training rules for IP tuning of gain and bias vectors derived in \cite{Schrauwen} are:

\begin{equation}
     \Delta a=\eta/a+\Delta bx 
     \label{eqda}
\end{equation}

\begin{equation}
     \Delta b=-\eta (-\mu/{\sigma ^2}+r/{\sigma ^2}(2{\sigma ^2}+1-r^2+\mu r))
     \label{eqdb}
\end{equation}

Here $\eta$ is learning rate and $x=W^{in}u+W^{res}r$ is the net input to the reservoir neurons. The Algorithm \ref{algorithmIP} presents the IP tuning procedure. All original vectors of features $f(i), i=1 \div n_f$ are fed consecutively to the ESN reservoir with zero initial state $r(0)=0$ for $n_{IP}$ IP tuning iterations (usually 3-5 are enough as in \cite{Schrauwen}). In present research, like in \cite{KopTontchev} the IP tuned reservoir neurons equilibrium states $r_e$ are exploited. The Algorithm \ref{algorithmeq} present the features extraction procedure.

\begin{algorithm}
\caption{IP tuning\label{algorithmIP}}
\begin{algorithmic}[2]
\STATE $IP~ tuning~ of~ ESN$
\STATE $Generate~ initial~ ESN~ reservoir$
\STATE $r(0)=0$
\STATE $a(0)=1$
\STATE $b(0)=0$
\FOR{$k=1\div n_{IP}$}
\FOR{$i=1\div n_{f}$}
\STATE $x(i)=W^{in}f(i)+W^{res}r(i-1)$ 
\STATE $r(i)=tanh(a(i-1)x(i)+b(i-1))$ 
\STATE $a(i)=a(i-1)+\Delta a(i)$
\STATE $b(i)=b(i-1)+\Delta b(i)$
\ENDFOR
\ENDFOR
\end{algorithmic}
\end{algorithm}

\begin{algorithm}
\caption{Features extraction\label{algorithmeq}}
\begin{algorithmic}[2]
\STATE $Extract~ features~ from~ IP~ tuned~ ESN$
\FOR{$i=1\div n_{f}$}
\STATE $r(0)=0$
\FOR{$k=1\div n_{it}$}
\STATE $x(k)=W^{in}f(i)+W^{res}r(k-1)$ 
\STATE $r(k)=tanh(ax(k)+b)$  
\ENDFOR
\STATE $r_e^i=r(k)$
\ENDFOR
\end{algorithmic}
\end{algorithm}

All original vectors of features $f(i), i=1 \div n_f$ are fed consecutively $n_{it}$ times as constant input $u(k)=f(i)=const., k=1 \div n_{it}$ to the ESN reservoir with zero initial state $r(0)=0$ until it settles to a new equilibrium state $r_e(f(i))=r(n_{it})$. $n_{it}$ is number of iterations needed to achieve steady state $r_e=r(k)=r(k-1)$ of the reservoir. Achieved in this way equilibrium states of the ESN reservoir $r_e^i$ for each original features vector $f(i)$ are considered as new features. 

In \cite{KopTontchev} it was investigated how the IP tuning of a randomly generated ESN reservoir led to clearer separation of the original multidimensional data after its projection to low dimensional space. The effect of IP tuning on reservoir equilibrium state was further investigated in \cite{KopIJCNN} demonstrating that it increases equilibrium states memory capacity and is strongly influenced by the original data structure. 

Provoked by the needs of gray scale images clustering a modified approach for features extraction was proposed. In contrast to \cite{KopTontchev}, here from each single feature per pixel (that is its intensity value $f(i)=pi(i), i=1 \div n_{pi}$) multiple features corresponding to ESN reservoir neurons equilibrium states $r_e^i$ were obtained. Here $n_f=n_{pi}$ is the number of the gray image pixels; $pi(i)$ is $i-th$ pixel intensity; The obtained in this way multidimensional feature vector $r_e^i$ for each pixel $pi(i)$ is subject to clustering. 

\section{Results and discussion}
\label{Results}
In order to investigate effects of proposed features extraction approach a benchmark image Lena was used. The original colour image was converted to gray scale and pixels intensities $pi$ were scaled in range $[-1,1]$. Next, all pixels intensities were applied to tune the gain and bias parameters of a randomly generated fully connected ESN reservoir with size $n_r=10$ and spectral radius $0.9$. The target Gaussian distribution of IP tuning was with zero mean and variance $\sigma=0.1$ and number of IP tuning iterations was set to $n_{IP}=5$. The number of iterations needed to achieve reservoir steady state was estimated to $n_{it}=50$. 

\begin{figure*}
\centering
\includegraphics[width=0.7\textwidth]{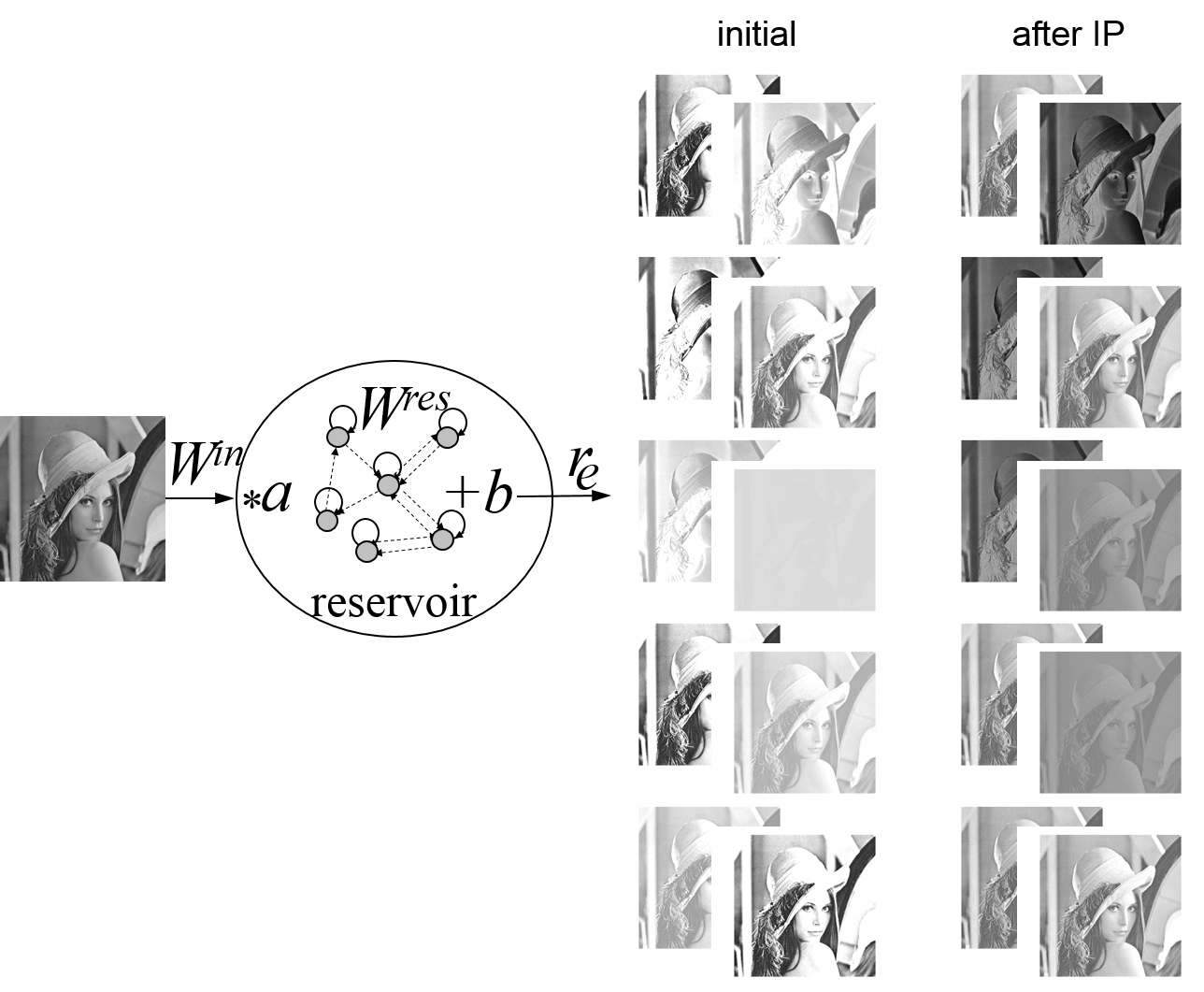}
\caption{Gray image features extraction.}
\label{ESNgray}       
\end{figure*}

\begin{figure}
\centering
\includegraphics[width=0.4\textwidth]{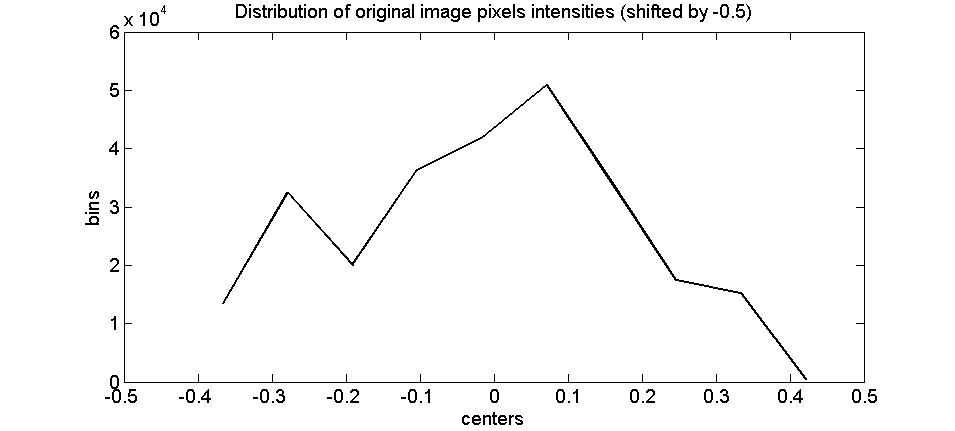} 
\includegraphics[width=0.4\textwidth]{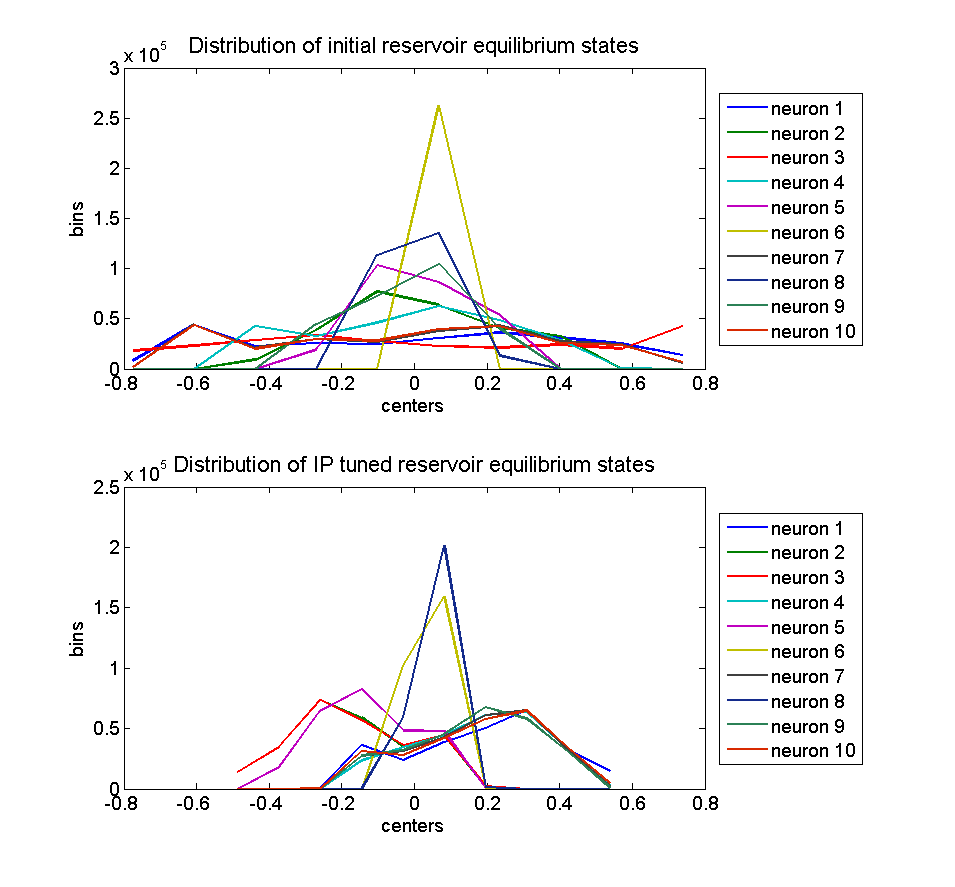}
\caption{Histogram of the original image pixels intensities shifted by $-0.5$ (top) vs. equilibrium states distributions of initial (middle) and IP tuned reservoir (bottom).}
\label{figESNhis}
\end{figure}

For the seek of comparison features were extracted by initial and IP tuned reservoir as shown on Fig. \ref{ESNgray}. Fig. \ref{figESNhis} shows histograms of the original image pixels intensities versus those of features extracted by initial and IP tuned ESN reservoirs. The observed effects of IP tuning are following: the reservoir equilibrium states were squeezed in narrow interval; the distribution of new equilibrium states reflects the original data distribution.

Next kmeans clustering algorithm was applied to separate image pixels into clusters using all original and all extracted by initial and IP tuned reservoir features. From Fig. \ref{figESNhis} is clear that features extracted from IP tuned reservoir can be visually separated into three groups - having maximum around $-0.3$, $0.1$ and $0.3$ respectively that is in accordance other segmentation approaches tested on Lena image. So the number of clusters was set to three here. The representative features from the three groups in Fig. \ref{figESNhis} (neurons 1, 3 and 8) were selected for clustering too. 

\begin{figure*}
\centering
\includegraphics[width=0.14\textwidth]{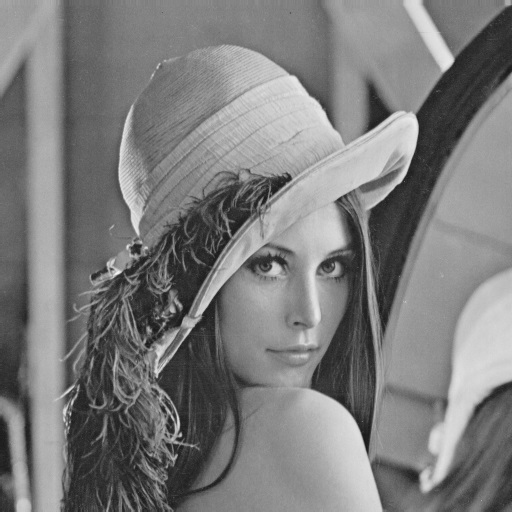} \\
\includegraphics[width=0.19\textwidth]{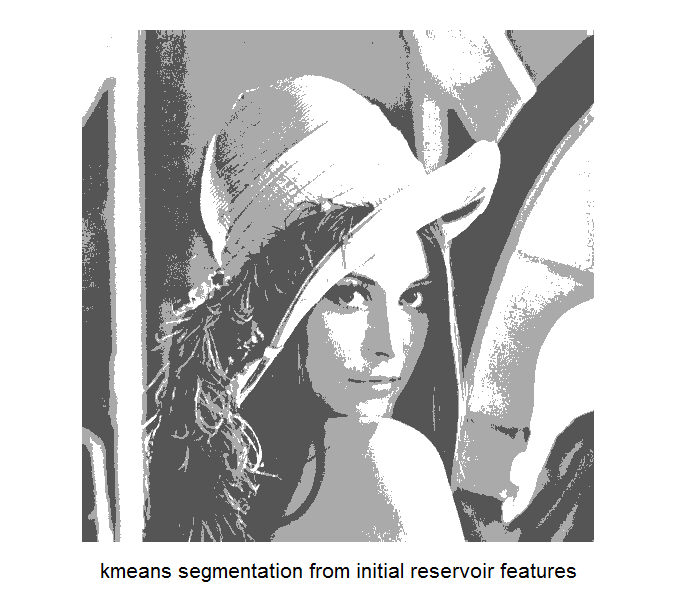}
\includegraphics[width=0.19\textwidth]{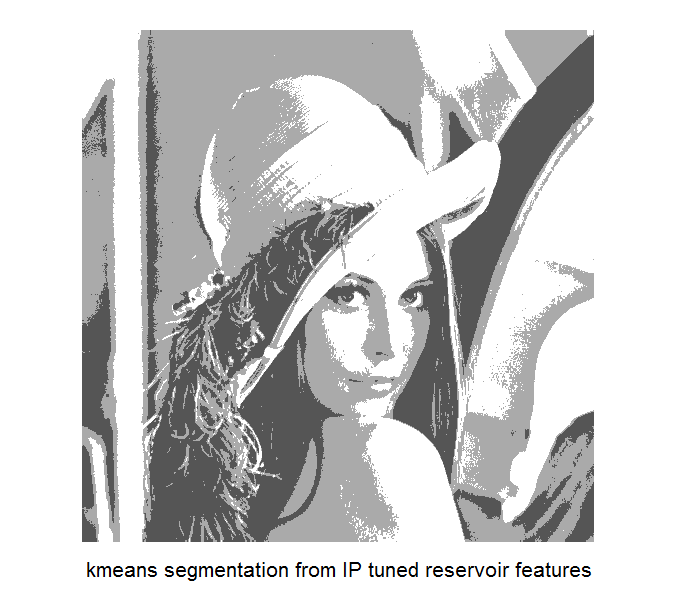}
\includegraphics[width=0.19\textwidth]{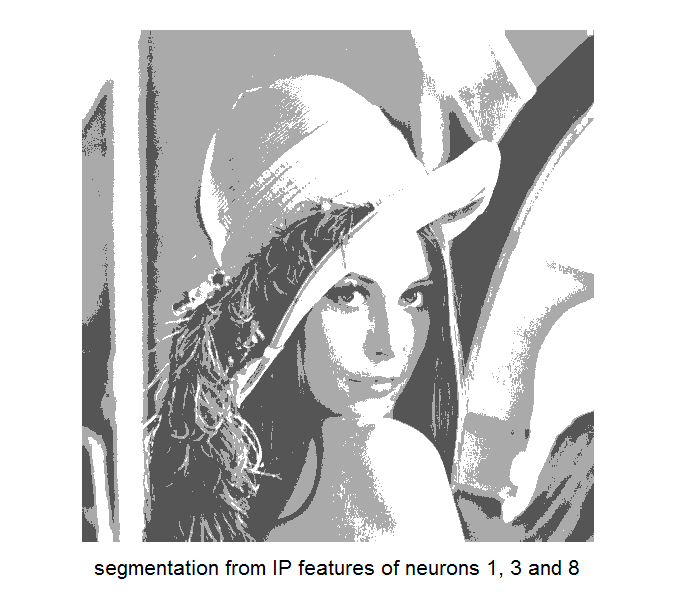}
\caption{Segmentation of the original gray image (on the top) into three clusters by kmeans clustering. Second row, from left to right, represents the clustering results achieved using all features extracted by the initial reservoir (left), by the IP tuned reservoir (middle) and by it representative neurons 1, 3 and 8 (right).}
\label{figLenaall_new}
\end{figure*}

Comparison of the original gray image and its segmentation via kmeans clustering are shown in Fig. \ref{figLenaall_new}. While segmentation using features extracted by initial random ESN reservoir looks blurred, the results achieved using IP tuned reservoir reveal sharper discrimination between image regions. Segmentation via representative features (neurons 1, 3 and 8) seems quite similar to that achieved using all features extracted by IP tuned reservoir. Looking at Fig. \ref{figLenaall_new}, it is observed that the results achieved using features extracted by IP tuned reservoir look quite similar to those obtained by kmeans clustering of original image pixels intensities. However, IP tunned reservoir features revealed a little bit more details from the original image, e.g. shades on the shoulder, contours in the background stuff etc. 

\begin{figure*}
\centering
\includegraphics[width=0.14\textwidth]{lena_gray.jpg} \\
\includegraphics[width=0.19\textwidth]{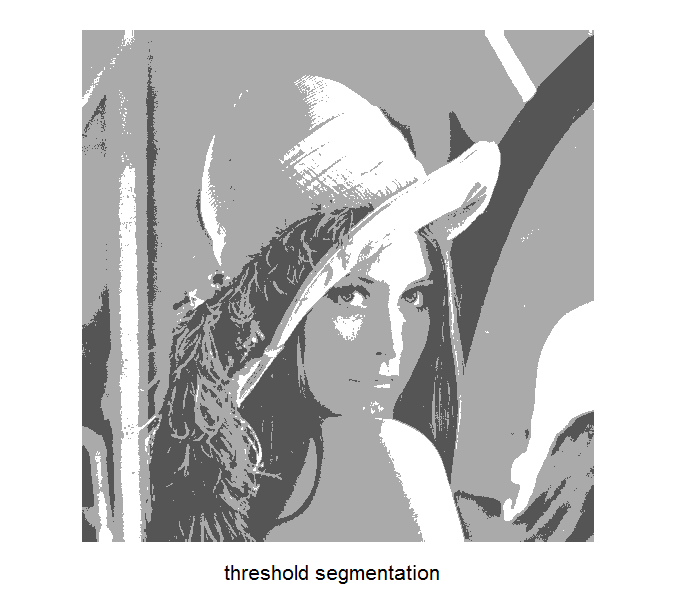}
\includegraphics[width=0.19\textwidth]{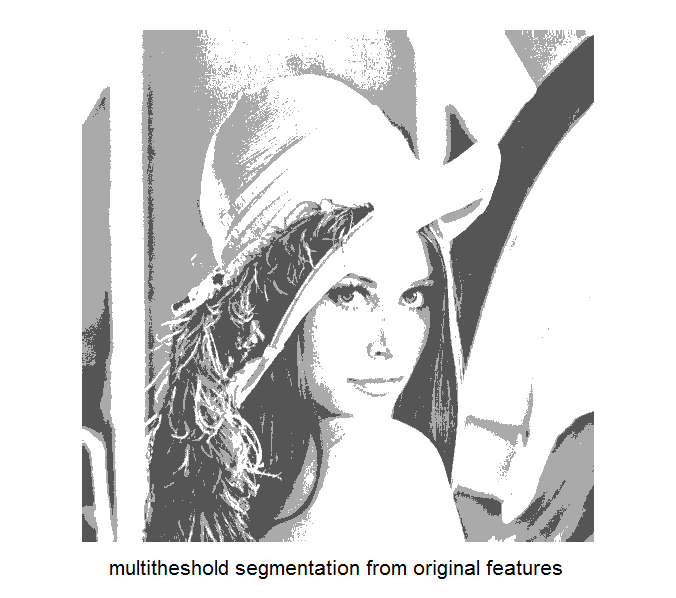} 
\includegraphics[width=0.19\textwidth]{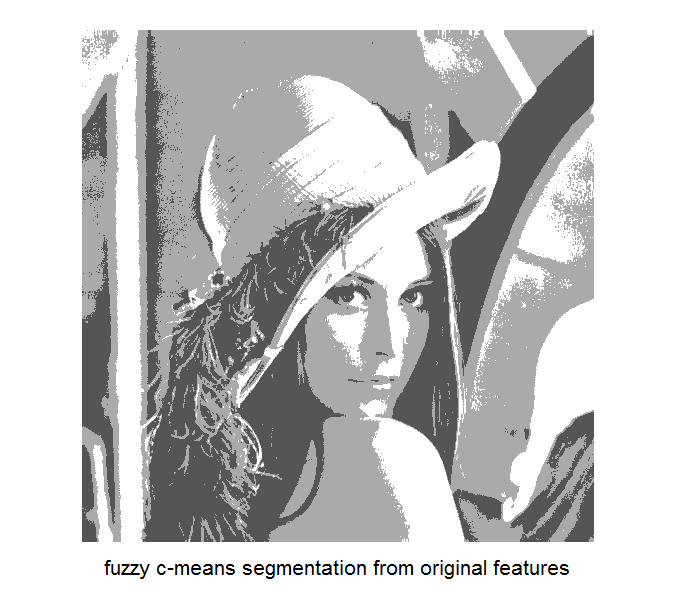} 
\includegraphics[width=0.19\textwidth]{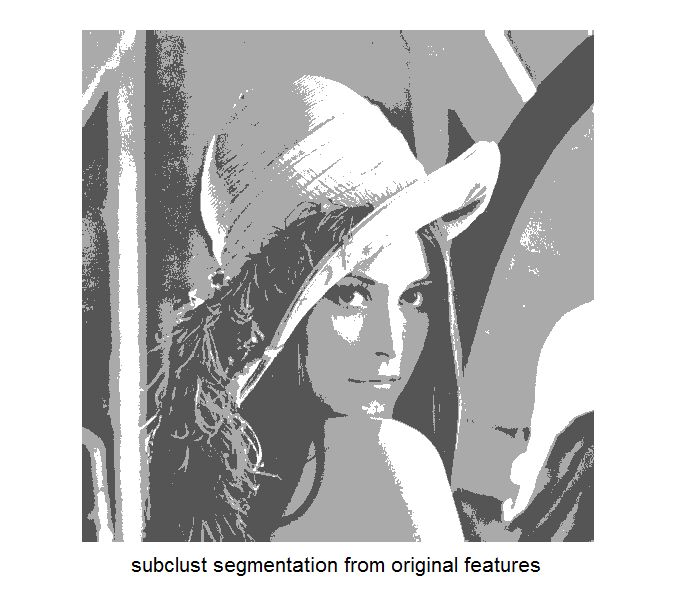}
\includegraphics[width=0.19\textwidth]{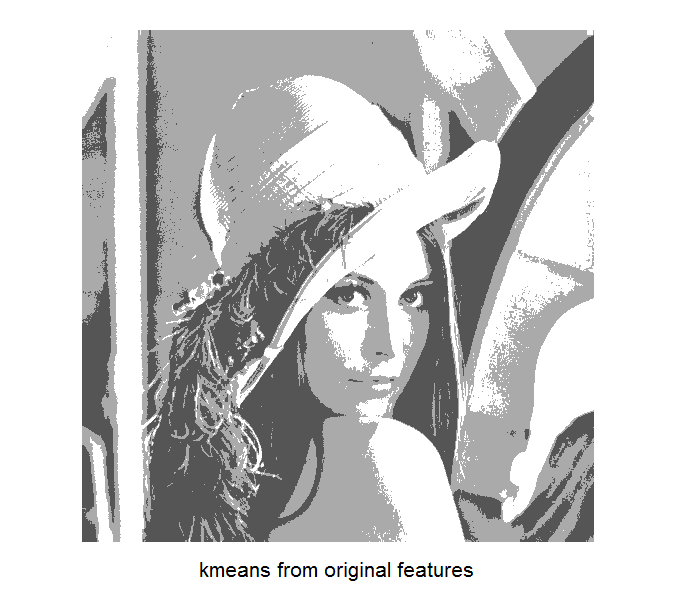} 
\caption{Segmentation of the original gray image (on the top) into three clusters using original image pixels intensities. Second row from left to right, represents the clustering results achieved via hard thresholding, multi-level threscholding, fuzzy c-means, subtractive clustering and kmeans clustering.}
\label{figLenaall}
\end{figure*}

For the seek of comparison the gray image was segmented using original pixels intensities and several clustering approaches: kmeans, hard thresholding using fixed threshold levels equally distributed within range of pixels intensities, multi-level thresholding using Otsu's method \cite{Otsu}, fuzzy c-means clustering and subtractive clustering \cite{Yager}. Fig. \ref{figLenaall} presents comparison of the original gray image and its segmentation by enumerated clustering approaches. It is obvious that kmeans segmentation outperforms all other approaches achieving sharper segmentation of the image. 

\begin{figure*}
\centering
\includegraphics[width=0.14\textwidth]{lena_gray.jpg} \\
\includegraphics[width=0.19\textwidth]{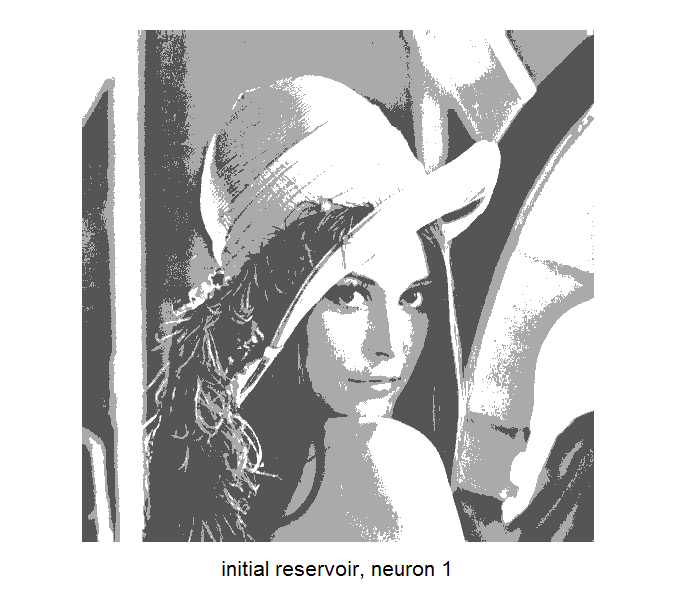}
\includegraphics[width=0.19\textwidth]{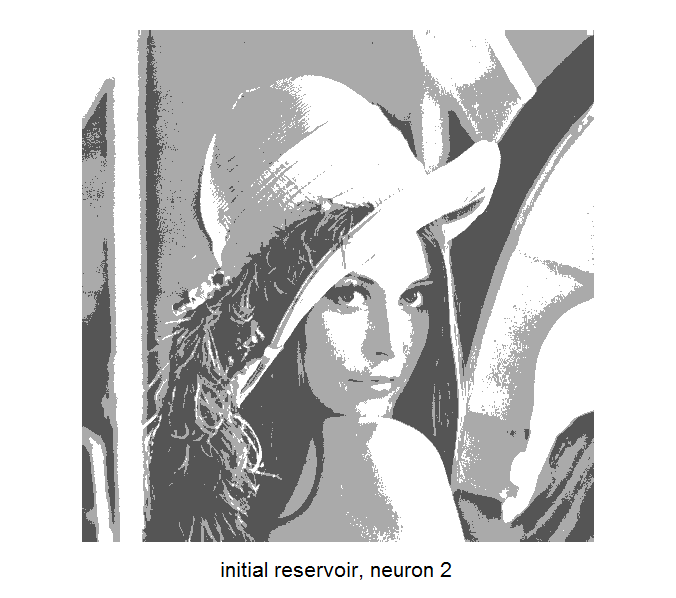} 
\includegraphics[width=0.19\textwidth]{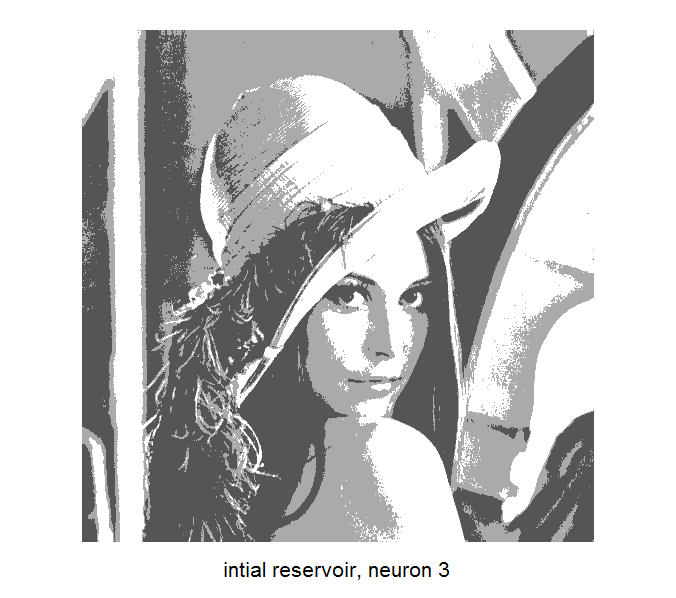} 
\includegraphics[width=0.19\textwidth]{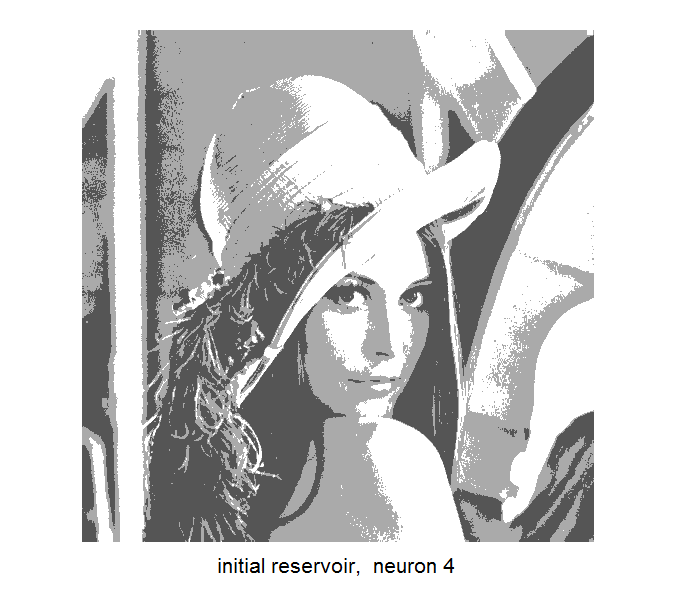} 
\includegraphics[width=0.19\textwidth]{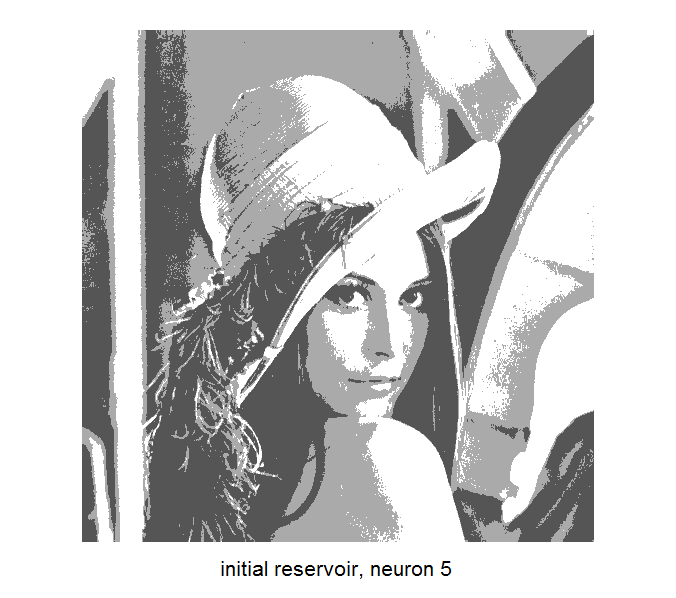}
\includegraphics[width=0.19\textwidth]{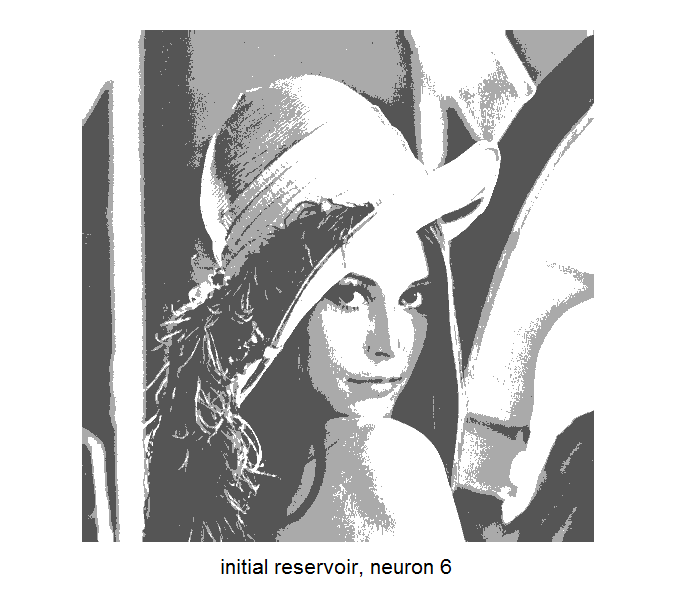} 
\includegraphics[width=0.19\textwidth]{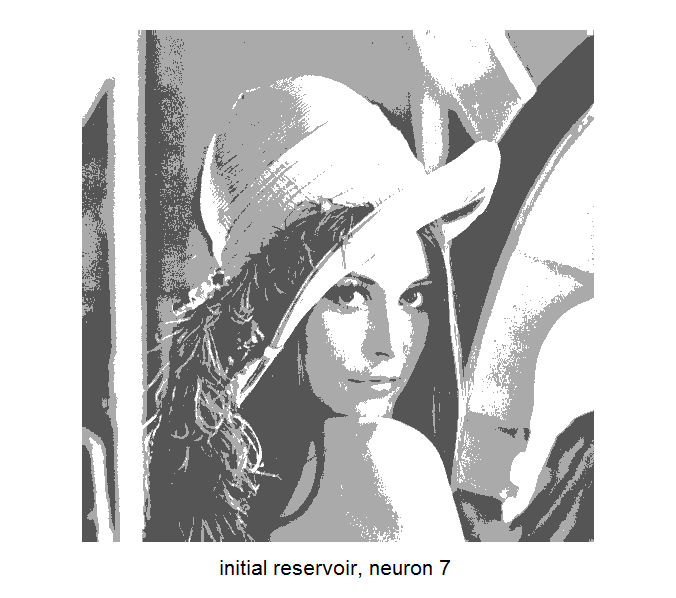}
\includegraphics[width=0.19\textwidth]{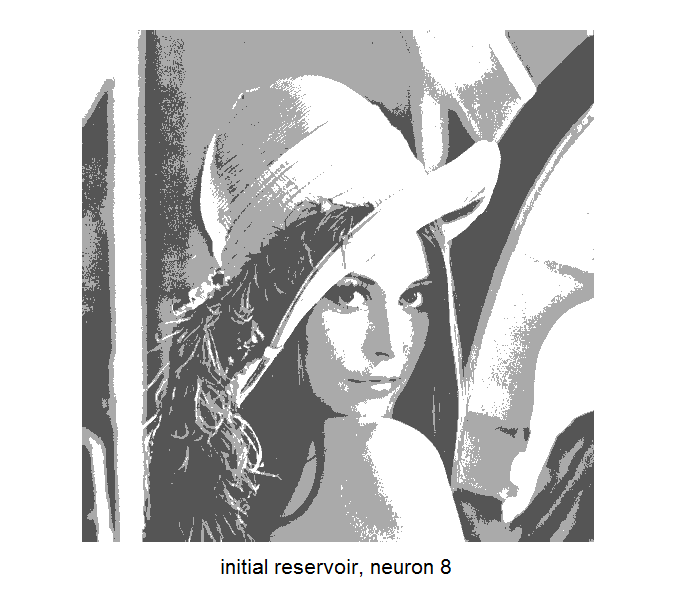} 
\includegraphics[width=0.19\textwidth]{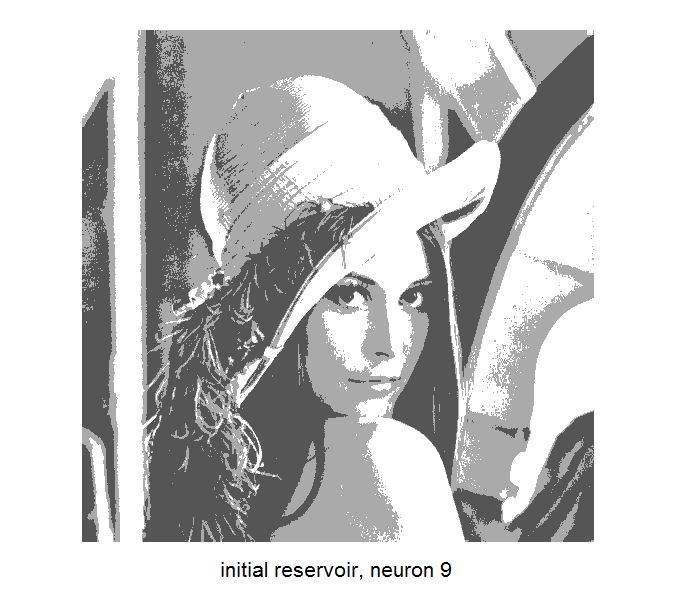} 
\includegraphics[width=0.19\textwidth]{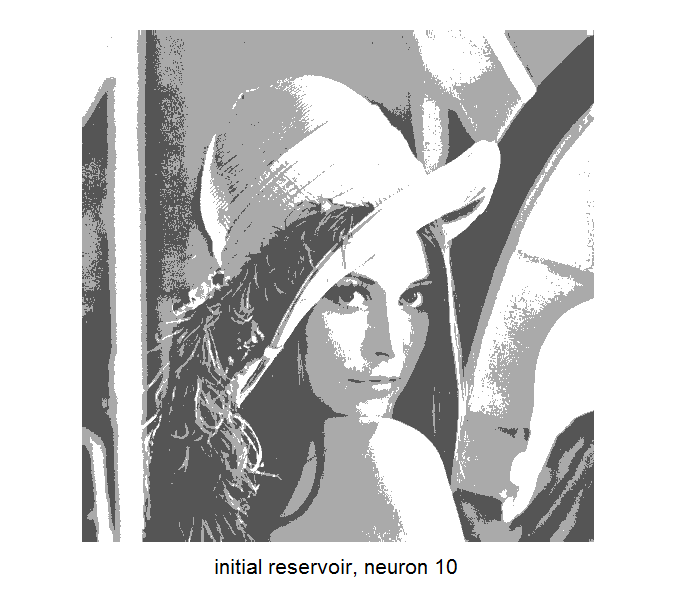}
\caption{Gray Lena image (on the top) and its segmentation from individual features extracted by initial reservoir.}
\label{figLenaini}
\end{figure*}

\begin{figure*}
\centering
\includegraphics[width=0.14\textwidth]{lena_gray.jpg} \\
\includegraphics[width=0.19\textwidth]{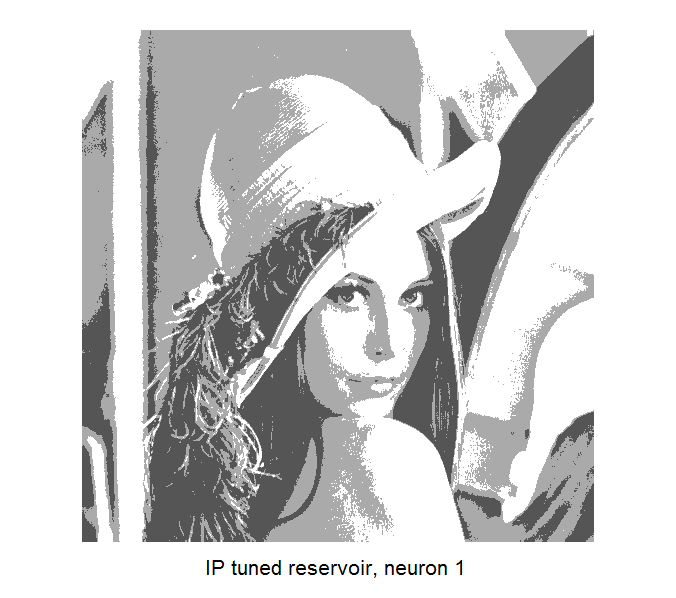}
\includegraphics[width=0.19\textwidth]{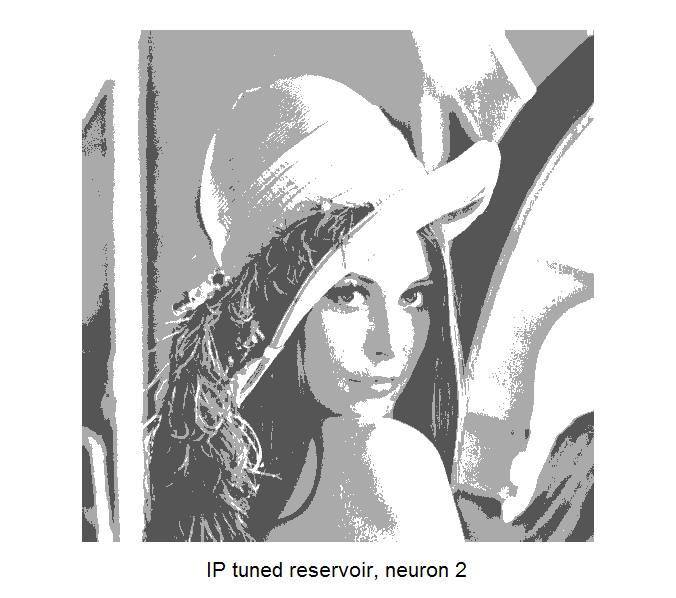}
\includegraphics[width=0.19\textwidth]{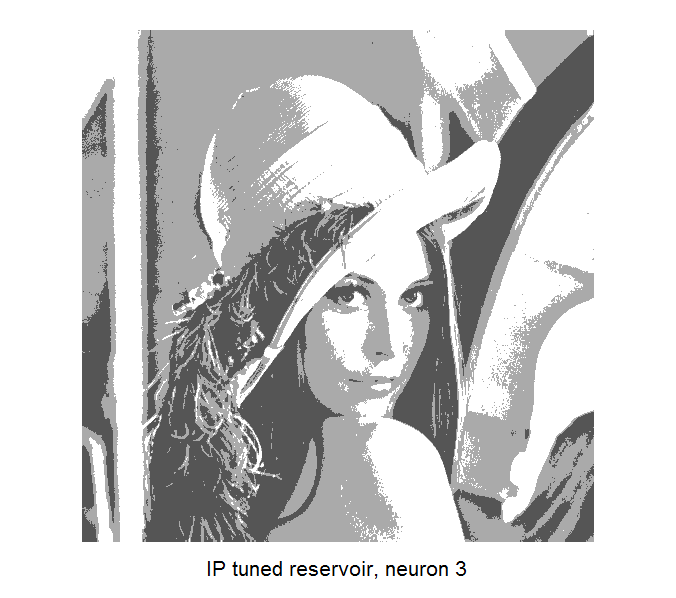} 
\includegraphics[width=0.19\textwidth]{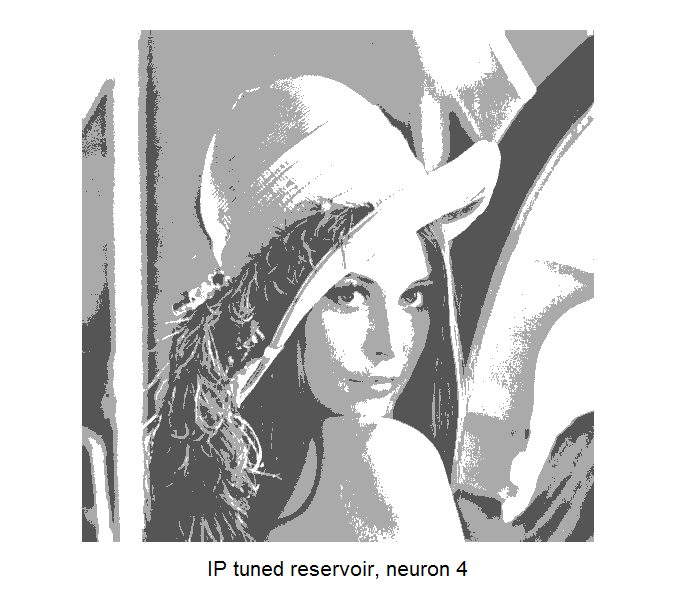}
\includegraphics[width=0.19\textwidth]{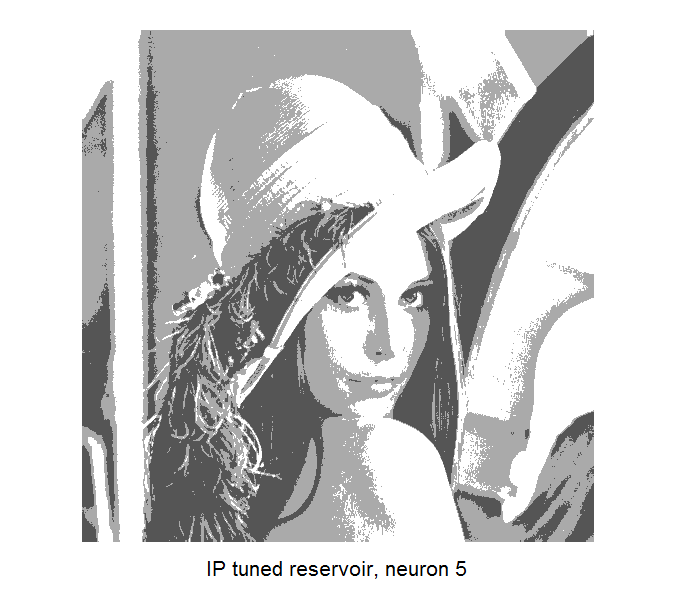}
\includegraphics[width=0.19\textwidth]{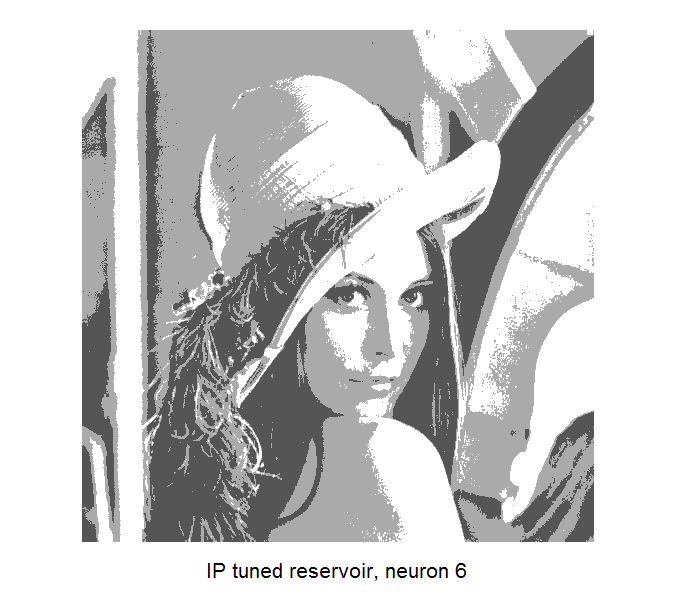} 
\includegraphics[width=0.19\textwidth]{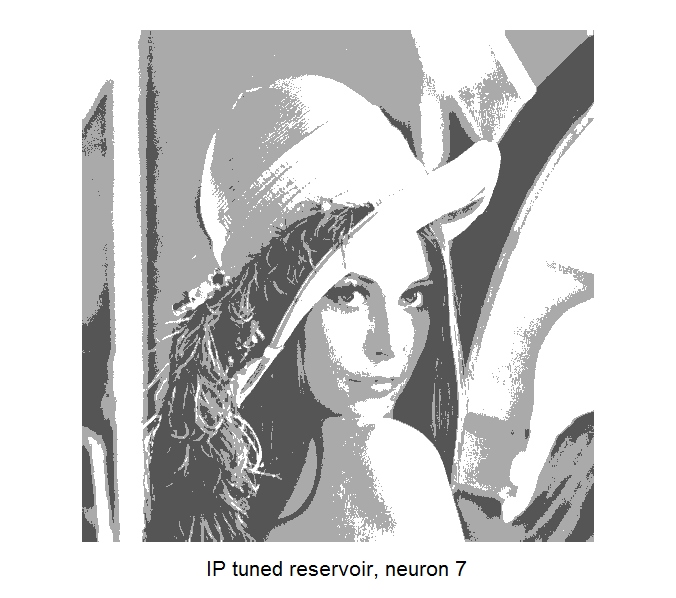}
\includegraphics[width=0.19\textwidth]{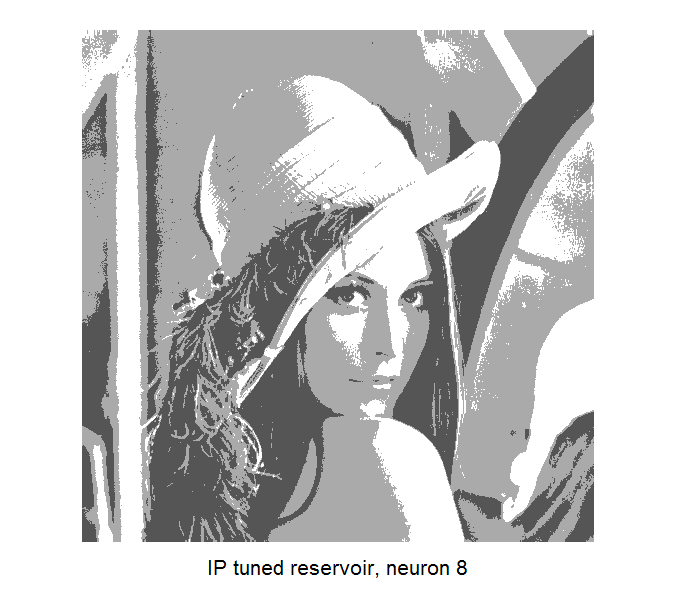}
\includegraphics[width=0.19\textwidth]{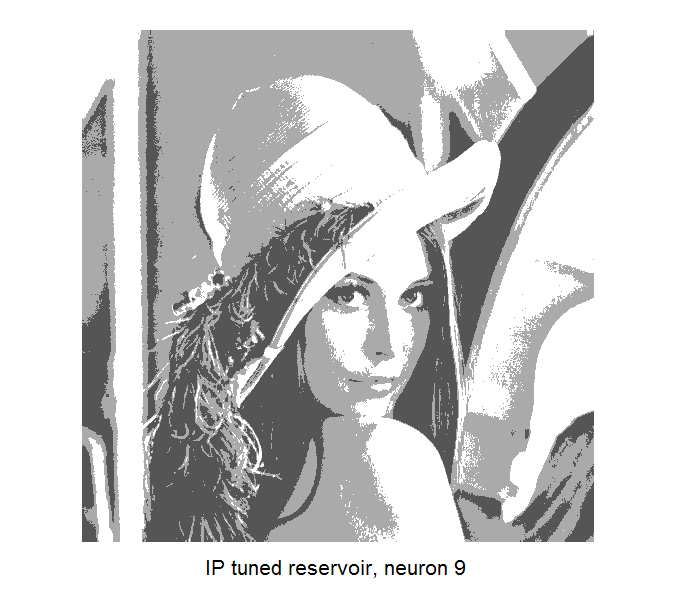} 
\includegraphics[width=0.19\textwidth]{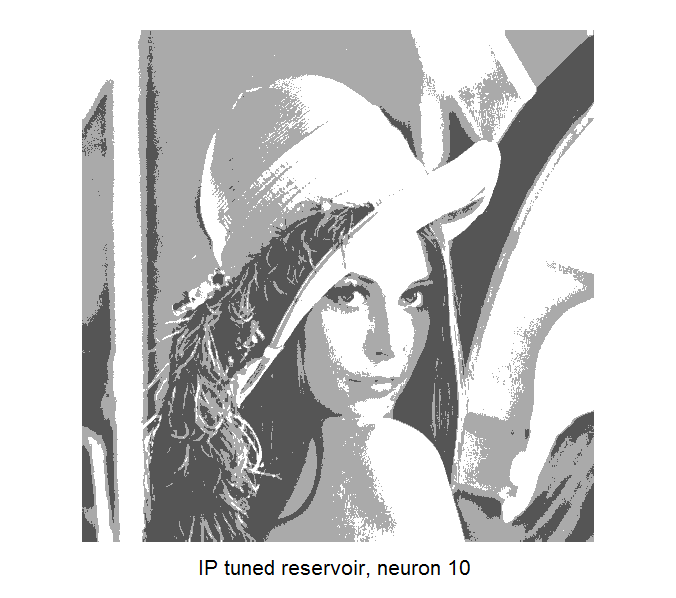}
\caption{Gray Lena image (on the top) and its segmentation from individual features extracted by IP tuned reservoir.}
\label{figLenaIP}
\end{figure*}

Fig. \ref{figLenaini} and \ref{figLenaIP} present results from kmeans clustering using individual features extracted by the initial and IP tuned reservoirs respectively.

Feature by feature comparison shows that IP tuning allows to achieve sharper discrimination between regions with different grades of pixels intensities. Besides, looking at individual features results, becomes clear that each neuron acts as a filter revealing different image characteristics. For example, neuron 8 of IP tuned reservoir reveals differences in pixels intensity at the left top corner of the original image that are not reveal by anyone of the rest of exploited features while it fails do distinguish shades at the right top corner of the image. 

Looking at Fig. \ref{figLenaini} and \ref{figLenaIP} leads to the conclusion that reservoir equilibrium states can be considered as image projections from different view points in the space. Thus in dependence on some additional, e.g. expert information, a proper for a given aim projection can be chosen for image segmentation. Close comparison of segmentation results from the original and IP tuned reservoir features however is in favour of the IP tuning approach since it reveals sharper discrimination of shapes in some regions of the image, e.g. some light bunches in the hair, shades on the shoulder, contours in the background stuff etc.

\section{Conclusions} 
\label{Concl}
The obtained results demonstrated that the proposed approach for gray images features extraction allows to reveal various hidden features discovering different view points to the image. They can be exploited in dependence on the specific needs and opinions of the experts in particular area interested in given image segmentation.  

These preliminary results are good basis for further development of hierarchical (deep) approach for gray images segmentation combining ESN and various clustering approaches.

Further investigations with different kind of gray images, e.g. X-ray photos, tomography images etc. can reveal various directions of application of the proposed approach. 

Combination of blind clustering with expert knowledge could help to further refine the approach adjusting it to specific expertise needs. A prospective direction of future work is also development of a support system for experts that will be able to choose the proper image projection among the automatically generated numerous projections generated from features extracted by IP tuned reservoir.








\end{document}